# AMHARIC-ARABIC NEURAL MACHINE TRANSLATION

Ibrahim Gashaw[1] and H L Shashirekha[2]

[1]Mangalore University, Department of Computer Science, Mangalagangotri, Mangalore-574199
ibrahimug1@gmail.com
[2]Mangalore University, Department of Computer Science, Mangalagangotri
hlsrekha@gmail.com

## ABSTRACT

*Many automatic translation works have been addressed between major European language pairs, by taking advantage of large scale parallel corpora, but very few research works are conducted on the Amharic-Arabic language pair due to its parallel data scarcity. Two Long Short-Term Memory (LSTM) and Gated Recurrent Units (GRU) based Neural Machine Translation (NMT) models are developed using Attention-based Encoder-Decoder architecture which is adapted from the open-source OpenNMT system. In order to perform the experiment, a small parallel Quranic text corpus is constructed by modifying the existing monolingual Arabic text and its equivalent translation of Amharic language text corpora available on Tanzile. LSTM and GRU based NMT models and Google Translation system are compared and found that LSTM based OpenNMT outperforms GRU based OpenNMT and Google Translation system, with a BLEU score of 12%, 11%, and 6% respectively.*

## 1. INTRODUCTION

"Computational linguistics from a computational perspective is concerned with understanding written and spoken language, and building artifacts that usually process and produce language, either in bulk or in a dialogue setting." [1].

Machine Translation (MT), the task of translating texts from one natural language to another natural language automatically, is an important application of Computational Linguistics (CL) and Natural Language Processing (NLP). The overall process of invention, innovation, and diffusion of technology related to language translation drive the increasing rate of the MT industry rapidly [2]. The number of Language Service Provider (LSP) companies offering varying degrees of translation, interpretation, localization, language, and social coaching solutions are rising in accordance with the MT industry [2]. Today many applications such as Google Translate and Microsoft Translator are available online for language translations.

There are different types of MT approaches, and many researchers have classified them in different ways. Oladosu et al. [3] have classified MT into two main categories: single and hybrid approaches. The single approach uses only one method to translate from one natural language to another, and it includes rule-based, direct, corpus-based, and knowledge-based approaches to MT. A hybrid approach is a combination of the statistical method and the rule-based approach, which includes a word-based model, a phrase-based model, a syntax-based model, and forest-based model. Single approaches achieve low performance because they are inconsistent and inflexible for large scale applications and give a shallower representation of knowledge resulting in lower quality and fluency of the output. The hybrid approach combines the strength of two or more approaches to improve the overall quality of MT [3].

Deep learning NMT approach is a recent approach of MT that produces high-quality translation results based on a massive amount of aligned parallel text corpora in both the source and target languages. Deep learning is part of a broader family of ML methods based on artificial neural networks [4]. It allows computational models that are composed of multiple processing layers to learn representations of data with various levels of abstraction. These methods have improved the state-of-the-art research in language translation [5]. NMT is one of the deep learning end-to-end learning approaches to MT that uses a large artificial neural network to predict the likelihood of a sequence of words, typically modeling entire sentences in a single integrated model. The advantage of this approach is that a single system can be trained directly on the source and target text no longer requiring the pipeline of specialized systems used in statistical MT. Many companies, such as Google, Facebook, and Microsoft, are already using NMT technology [6]. NMT has recently shown promising results on multiple language pairs. Nowadays, it is widely used to solve translation problems for many language pairs. However, much of the research on this area has focused on European languages despite these languages being very rich in resources.

MT is a very ambitious research task in NLP, and the demand for it is growing. Several MT systems have been developed all over the world, particularly from English to other natural languages, such as Arabic, Germany, Chinese, French, Hindi, Japanese, Spanish, and Urdu [7]. Though a limited amount of work has been done in different Ethiopian languages in the field of NLP, the MT system for Amharic-Arabic language pair is still in its infancy due to lack of parallel corpora. Therefore, it is essential to construct Amharic-Arabic parallel text corpora, which is very much required to develop the Amharic to Arabic NMT system.

Amharic language is the national language of Ethiopia spoken by 26.9% of Ethiopia's population as mother tongue and spoken by many people in Israel, Egypt, and Sweden. Arabic is a Semitic language spoken by 250 million people in 21 countries as the first language and serving as a second language in some Islamic countries. Ethiopia is one of the nations, which have more than 33.3% of the population who follow Islam, and they use the Arabic language to teach religion and for communication purposes. Both of these languages belong to the Semitic family of languages, where the words are formed by modifying the root itself internally and not merely by the concatenation of affixes to word roots [8].

NMT has many challenges, such as; domain mismatch, size of training data, rare words, long sentences, word alignment, and beam search [9] depending on the language pairs. Some of these challenges are addressed in this paper.

## 2. RELATED WORK

Different attempts have been made in the past to construct parallel text corpus from Amharic and Arabic languages pair with the English language. Some of these works are as follows;

G.A Mekonnen, A. Nurnberger, and Bat [10], describe the acquisition, preprocessing, segmentation, and alignment of Amharic-English parallel corpus that consists of 1,45,820 Amharic-English parallel sentences (segments) from various sources. This corpus is larger in size than previously compiled Amharic-English Bilingual Corpus (AEBC), which is hosted by

European Language Resource Association with the size of 13,379 aligned segments (sentences) and the Low Resource Languages for Emergent Incidents (LORELEI) developed by Strassel and Tracey [11], that contains 60,884 segments.

Sakre et al., [12], presents a technique that aimed to construct an Arabic-English corpus automatically through web mining. The system crawled the host using GNU Wget in order to obtain English and Arabic web pages and then created candidate parallel pairs of documents by filtering them according to their similarity in the path, file name, creation date, and length. Finally, the technique measured the parallelism similarity between the candidate pairs according to the number of transnational tokens found between an English paragraph and its three Arabic neighbor paragraphs. However, in this work, they did not test or compare different models of statistical translation using the constructed parallel corpus.

Ahmad et al. [13] reported the construction of one million words English-Arabic Political Parallel Corpus (EAPPC) that consists of 351 English and Arabic original documents and their translations. The texts were meta-annotated, segmented, tokenized, English-Arabic aligned, stemmed, and POS-tagged. Arabic and English data were extracted from the official website of 'His Majesty King Abdullah II' and from 'His book' and then reprocessed from metadata annotation to alignment. They built a parallel concordancer, which consisted of two parts: the application through which the end-user interacts with the corpus and the database which stores the parallel corpus. Experiments carried out examined the translation strategies used in rendering a culture-specific term, and results demonstrated the ease with which knowledge about translation strategies can be gained from this parallel corpus.

Inoue et al. [14], describe the creation of Arabic-Japanese portion of a parallel corpus of translated news articles, which is collected at Tokyo University of Foreign Studies (TUFS). Part of the corpus is manually aligned at the sentence level for development and testing. The first results of Arabic-Japanese phrase-based MT trained on this corpus reported a BLEU score of 11.48.

Alotaibi [15], described an ongoing project at the College of Languages and Translation, King Saudi University, to compile a 10-million-word Arabic-English parallel corpus to be used as a resource for translation training and language teaching. The corpus has been manually verified at different stages, including translations, text segmentation, alignment, and file preparation, to enhance its quality. The corpus is available in XML format and through a user-friendly web interface, which includes a concordance that supports bilingual search queries and several filtering options.

All the above-related works use different lexical resources like machine-readable bilingual dictionaries or parallel corpora, for probability assessment or translation. However, when there is a lack of such a lexical resource, an alternative approach should be available [16]. Nowadays, NMT Models are widely used to solve various translation problems. Learning Phrase Representations using Recurrent Neural Network (RNN) Encoder-Decoder for statistical MT [17] benefits more natural language-related applications as it can capture the linguistic regularities in multiple word level as well as phrase level. But it is limited to target phrases, instead of using a phrase table.

Dzmitry et al. [18], extended NMT Encoder-Decoder that encodes a source sentence into a fixed-length vector, which is used by a decoder to generate a translation. It automatically search for relevant parts of a source sentence to predict a target word without having to form these parts like a hard segment explicitly. Their method yielded good results on longer sentences, and the alignment mechanisms are jointly trained towards a better log-probability of producing correct translations that need high computational cost.

A. Almahairi et al. [19], proposed NMT for the task of Arabic translation in both directions (Arabic-English and English-Arabic) and compared a Vanilla Attention-based NMT system against a Vanilla Phrase-based system. Preprocessing Arabic texts can increase the performance of the system, especially normalization, but the model consumes much time for training.

## 3. TRANSLATION CHALLENGES OF AMHARIC AND ARABIC LANGUAGES

Amharic and Arabic Languages are characterized by complex, productive morphology, with a basic word-formation mechanism, root-and-pattern. The root is a sequence of consonants, and the pattern is a sequence of Vowels (V) and Consonants (C) with open slots in it. It combines with the pattern through a process called interdigitating (intercalation): each letter of the root (radical) fills a slot in the pattern. For example, the Amharic root s.b.r (sabr) denoting a notion of breaking, combines with the pattern CVCC (the slots and vowels are denoted by C and V respectively) [20].

In addition to the unique root-and-pattern morphology, they are characterized by a productive system of more standard affixation processes. These include prefixes, suffixes, infixes, and circumfixes, which are involved in both inflectional and derivational processes. Consider the Arabic word "وسوف يكتوبنه" (wasawf yaktwubunahu) and its English translation "and they will write it". A possible analysis of these complex words defines the stem as "aktub" (write), with an inflectional circumfix, "y-uwna", denoting third person masculine plural, an inflectional suffix, "ha" (it), and two prefixes, "sa" (will) and "wa" (and).

Morphological analysis of words in a text is the first stage of most natural language applications. Morphological processes define the shape of words. They are usually classified into two types of processes [21];

1. A derivational process that deals with word-formation; such methods can create new words from existing ones, potentially changing the category of the original word. For example, from the Arabic root "كتب" (wrote), the following words are derived; "الكآتيب" (the writer), "الكيتآب" (the book), "المكتبة" (the library), "مكتبه" (library). The same is true in Amharic also. For example, from Amharic root "ገፈ" (he wrote), the following words are derived; "ፀፈ" (writer), "መፅሀፍ" (the book), "ቤተመፅሃፍ" (the library).

2. Inflectional processes are usually highly productive, applying to most members of a particular word class. For example, Amharic nouns inflect for number, so most nouns occur in two forms, singular (which is considered in the citation form) and plural, regularly obtained by adding the suffix "ዎች" to the base form. This process makes the translation ambiguous. The considerable number of potential types of words and the difficulty of handling out-of-lexicon items (in particular, proper names) combined with prefix or suffix makes the computation very challenging. For example in the word "aysäbramm" the prefix "ay" and the suffix "amm" (he doesn't break) are out-of-lexicon items.

The main lexical challenge in building NLP systems for Amharic and Arabic languages is the lack of machine-readable lexicons, which are vital resources. The absence of capitalization in Arabic and Amharic languages makes it hard to identify proper nouns, titles, acronyms, and abbreviations. Sentences in the Arabic language are usually long, and punctuation has no or little effect on the interpretation of the text.

Standard preprocessing techniques such as capitalization, annotation, and normalization cannot be performed on Amharic and Arabic languages due to issues of orthography. A single token in these languages can be a sequence of more than one lexical item, and hence be associated with a sequence of tags. For example, the Amharic word "አስፈረደችብኝ" ("asferedachibegn"), where translated to English will be "a case she initiated against me was decided in her favor". The word is built from the causative prefix "as" (causes), a perfect stem "ferede" (judged), a subject maker clitics "achi" (she), a benefactive marker "b" (against) and the object pronoun "egn" (I).

Contextual analysis is essential in both languages to understand the exact meaning of some words. For example, in Amharic, the word "ገና" can have the meaning of "Christmas holiday" or "waiting for something until it happens." Diacritics (vowels) are most of the time, omitted from the Arabic text, which makes it hard to infer the word meaning and complex morphological rules should be used to tokenize and parse the text.

The corpus of the Arabic language has a bias towards religious terminology as a relatively high frequency of religious terms and phrases are found. Characters are sometimes stretched for justified text, which hinders the exact match for the same word. Synonyms are very common in Arabic. For example, "year" has three synonyms عـام, حـول, سنه, and all are widely used in everyday communication.

Discretization is defined as a symbol over and underscored letters, which are used to indicate the proper pronunciations as well as for disambiguation purposes. Its absence in Arabic texts poses a real challenge for Arabic NLP, as well as for translation, leading to high ambiguity. Though the use of discretization is significant for readability and understanding, they don't appear in most printed media in Arabic regions nor on Arabic Internet web sites. They are visible in the Quran, which is fully discretized to prevent misinterpretation [8].

## 4. CONSTRUCTION OF AMHARIC-ARABIC PARALLEL TEXT CORPUS

Construction of a parallel corpus is very challenging and needs a high cost of human expertise. MT can produce high-quality translation results based on a massive amount of aligned parallel text corpora in both the source and target languages [7]. MT systems need resources that can provide an interpretation/suggestion of the source text and a translation hypothesis. Parallel corpus consists of parallel text that can promptly locate all the occurrences of one language expression to another language expression and is one of the significant resources that could be utilized for MT tasks. [22].

As Amharic-Arabic parallel text corpora are not available for MT tasks, we have constructed a small parallel text corpus by modifying the existing monolingual Arabic and its equivalent translation of Amharic language text corpora available on Tanzile [23]. Quran text corpus consists of 114 chapters, 6236 verses, and 157935 words. The organization of the Quran text is categorized into verses (sequence of sentences and phrases). A sample verse in Arabic and its equivalent translation in Amharic and English is shown in Table 1. A total number of 13,501 Amharic-Arabic parallel sentences corpora have been constructed to train the Amharic to Arabic NMT system by splitting the verses manually into separate sentences of Amharic language as a source sentence and Arabic language as a target sentence as shown in Table 2. The total size of the corpus is 3.2MB, and it is split into training (80%) and test (20%).

The chapters in the Quran start with "بِسْمِ اللَّـهِ الرَّحْمَـٰنِ الرَّحِيمِ" (in the name of Allah the most gracious and the most merciful) in the Arabic text corpus. As it is merged with the first verses of each chapter, we split it into a separate line. In the case of Amharic corpus, the equivalent translation of this sentence is placed only in the first chapter. Therefore, it is added before the first line of the equivalent translated Amharic verses.

Table 1. Sample Verse of Quran in Arabic and its equivalent translation in Amharic and English

| Chapter No: Verse No | Original Arabic Verses | Equivalent translation of Amharic Verses | Equivalent translation of English Verses |
|---|---|---|---|
| 2:282 | يَا أَيُّهَا الَّذِينَ آمَنُوا إِذَا تَدَايَنتُم بِدَيْنٍ إِلَىٰ أَجَلٍ مُّسَمًّى فَاكْتُبُوهُ ۚ وَلْيَكْتُب بَّيْنَكُمْ كَاتِبٌ بِالْعَدْلِ ۚ وَلَا يَأْبَ كَاتِبٌ أَن يَكْتُبَ كَمَا عَلَّمَهُ اللَّـهُ ۚ فَلْيَكْتُبْ وَلْيُمْلِلِ الَّذِي عَلَيْهِ الْحَقُّ وَلْيَتَّقِ اللَّـهَ رَبَّهُ وَلَا يَبْخَسْ مِنْهُ شَيْئًا ۚ فَإِن كَانَ الَّذِي عَلَيْهِ الْحَقُّ سَفِيهًا أَوْ ضَعِيفًا أَوْ لَا يَسْتَطِيعُ أَن يُمِلَّ هُوَ فَلْيُمْلِلْ وَلِيُّهُ بِالْعَدْلِ ۚ وَاسْتَشْهِدُوا شَهِيدَيْنِ مِن رِّجَالِكُمْ ۖ فَإِن لَّمْ يَكُونَا رَجُلَيْنِ فَرَجُلٌ وَامْرَأَتَانِ مِمَّن تَرْضَوْنَ مِنَ الشُّهَدَاءِ أَن تَضِلَّ إِحْدَاهُمَا فَتُذَكِّرَ إِحْدَاهُمَا الْأُخْرَىٰ ۚ وَلَا يَأْبَ الشُّهَدَاءُ إِذَا مَا دُعُوا ۚ وَلَا تَسْأَمُوا أَن تَكْتُبُوهُ صَغِيرًا أَوْ كَبِيرًا إِلَىٰ أَجَلِهِ ۚ ذَٰلِكُمْ أَقْسَطُ عِندَ اللَّـهِ وَأَقْوَمُ لِلشَّهَادَةِ وَأَدْنَىٰ أَلَّا تَرْتَابُوا ۖ إِلَّا أَن تَكُونَ تِجَارَةً حَاضِرَةً تُدِيرُونَهَا بَيْنَكُمْ فَلَيْسَ عَلَيْكُمْ جُنَاحٌ أَلَّا تَكْتُبُوهَا ۗ وَأَشْهِدُوا إِذَا تَبَايَعْتُمْ ۚ وَلَا يُضَارَّ كَاتِبٌ وَلَا شَهِيدٌ ۚ وَإِن تَفْعَلُوا فَإِنَّهُ فُسُوقٌ بِكُمْ ۗ وَاتَّقُوا اللَّـهَ ۖ وَيُعَلِّمُكُمُ اللَّـهُ ۗ وَاللَّـهُ بِكُلِّ شَيْءٍ عَلِيمٌ | እናንተ ያመናችሁ ሆይ! እስከ ተወሰነ ጊዜ ድረስ በዕዳ በተዋዋላችሁ ጊዜ ጻፉት። ጸሐፊም በመካከላችሁ በትክክል ይጻፍ። ጸሐፊም አላህ እንዳሳወቀው መጻፍን እንቢ አይበል። ይጻፍም፤ ያም ዕዳው ያለበት ሰው በቃላ ያስጽፍ። አላህንም ጌታውን ይፍራ። ከእርሱም (ካለበት ዕዳ) ምንንም አያጉድል። ያም ዕዳው ያለበት ቂል ወይም ደካማ፣ ወይም በቃላ ማስጻፍን የማይችል ቢኾን ዋቢው በትክክል ያስጽፍለት። ከወንዶቻችሁም ሁለትን ምስክሮች አስመስክሩ። ሁለትም ወንዶች ባይኾኑ ከምስክሮች ሲኾኑ ከምትወዷቸው የኾኑ አንድ ወንድና አንደኛዋ ስትሳሳ አንደኛይቱ ሌላዋን ታስታውስ ዘንድ ሁለት ሴቶች (ይመስክሩ)። ምስክሮችም በተጠሩ ጊዜ እንቢ አይበሉ። (ዕዳው) ትንሽ ወይም ትልቅ ቢኾንም እስከ ጊዜው ድረስ የምትጽፉት ከመኾን አትሰልቹ። እንዲህ ማድረጋችሁ አላህ ዘንድ በጣም ትክክል ለምስክርነትም አረጋጋጭ ላለመጠራጠራችሁም በጣም ቅርብ ነው። ግን በመካከላችሁ እጅ በጅ የምትቀበበሲት ንግድ ብትኾን ባትጽፉዋት በናንተ ላይ ኃጢአት የለባችሁም። በተሻጣችሁም ጊዜ አስመስክሩ። ጸሐፊም ምስክርም (በላ ጉዳዩ ጋር) አይዳዳሱ። (ይህንን) ብትሠሩም እርሱ በእናንተ (የሚጠጋ) አመጽ ነው። አላህንም ፍሩ። አላህም ያሳውቃችኋል። አላህም ነገሩን ሁሉ ዐዋቂ ነው። | "O believers, when you negotiate a debt for a fixed term, draw up an agreement in writing, though better it would be to have a scribe write it faithfully down; and no scribe should refuse to write as God has taught him, and write what the borrower dictates, and have a fear of God, his Lord, and not leave out a thing. If the borrower is deficient of mind or infirm, or unable to explain, let the guardian explain judiciously; and have two of your men to act as witnesses; but if two men are not available, then a man and two women you approve, so that in case one of them is confused the other may prompt her. When the witnesses are summoned, they should not refuse (to come). But do not neglect to draw up a contract, big or small, with the time fixed for paying back the debt. This is more equitable in the eyes of God, and better as evidence and best for avoiding doubt. But if it is a deal about some merchandise requiring transaction face to face, there is no harm if no (contract is drawn up) in writing. Have witnesses to the deal (and make sure) that the scribe or the witness is not harmed. If he is, it would surely be sinful on your part. And have a fear of God, for God gives you the knowledge, and God is aware of everything." |

Table 2. Split Sentences of Quran Chapter 2 Verse number 282

| Original Arabic Sentences | Equivalent translation of Amharic sentences | Equivalent translation of English sentences |
|---|---|---|
| يَا أَيُّهَا الَّذِينَ آمَنُوا | እናንተ ያመናችሁ ሆይ | O believers |
| إِذَا تَدَايَنتُم بِدَيْنٍ إِلَىٰ أَجَلٍ مُّسَمًّى فَاكْتُبُوهُ | እስከ ተወሰነ ጊዜ ድረስ በዕዳ በተዋዋላችሁ ጊዜ ጻፉት | when you negotiate a debt for a fixed term, draw up an agreement in writing |
| وَلْيَكْتُب بَّيْنَكُمْ كَاتِبٌ بِالْعَدْلِ | ጸሐፊም በመካከላችሁ በትክክል ይጻፍ | though better it would be to have a scribe write it faithfully down |
| وَلَا يَأْبَ كَاتِبٌ أَن يَكْتُبَ كَمَا عَلَّمَهُ اللَّهُ | ጸሐፊም አላህ እንዳሳወቀው መጻፍን እንቢ አይበል | and no scribe should refuse to write as God has taught him |
| فَلْيَكْتُبْ | ይጻፍም | and write |
| وَلْيُمْلِلِ الَّذِي عَلَيْهِ الْحَقُّ | ያም በርሱ ላይ ዕዳው ያለበት ሰው በቃሉ ያስጽፍ | what the borrower dictates |
| وَلْيَتَّقِ اللَّهَ رَبَّهُ | አላህንም ጌታውን ይፍራ | and have fear of God, his Lord |
| وَلَا يَبْخَسْ مِنْهُ شَيْئًا | ከእርሱም ካለበት ዕዳ ምንንም አያጉድል | and not leave out a thing |
| فَإِن كَانَ الَّذِي عَلَيْهِ الْحَقُّ سَفِيهًا أَوْ ضَعِيفًا أَوْ لَا يَسْتَطِيعُ أَن يُمِلَّ هُوَ فَلْيُمْلِلْ وَلِيُّهُ بِالْعَدْلِ | ያም በርሱ ላይ ዕዳው ያለበት ቂል ወይም ደካማ ወይም በቃሉ ማስጻፍን የማይችል ቢኾን ዋቢው በትክክል ያስጽፍለት | If the borrower is deficient of mind or infirm, or unable to explain, let the guardian explain judiciously |
| وَاسْتَشْهِدُوا شَهِيدَيْنِ مِن رِّجَالِكُمْ | ከወንዶቻችሁም ሁለትን ምስክሮች አስመስክሩ | and have two of your men to act as witnesses |
| فَإِن لَّمْ يَكُونَا رَجُلَيْنِ فَرَجُلٌ وَامْرَأَتَانِ مِمَّن تَرْضَوْنَ مِنَ الشُّهَدَاءِ أَن تَضِلَّ إِحْدَاهُمَا فَتُذَكِّرَ إِحْدَاهُمَا الْأُخْرَىٰ | ሁለትም ወንዶች ባይኾኑ ከምስክሮች ሲኾኑ ከምትወዳቸው የኾኑ አንድ ወንድና አንደኛዋ ስትሳሳት አንደኛይቱ ሌላዋን ታስታውሳት ዘንድ ሁለት ሴቶች ይመስክሩ | but if two men are not available, then a man and two women you approve, so that in case one of them is confused the other may prompt her |
| وَلَا يَأْبَ الشُّهَدَاءُ إِذَا مَا دُعُوا | ምስክሮችም በተጠሩ ጊዜ እንቢ አይሉ | When the witnesses are summoned they should not refuse (to come). |
| وَلَا تَسْأَمُوا أَن تَكْتُبُوهُ صَغِيرًا أَوْ كَبِيرًا إِلَىٰ أَجَلِهِ | ዕዳው ትንሽ ወይም ትልቅ ቢኾንም እስከ ጊዜው ድረስ የምትጽፉት ከመኾን አትሰልቹ | But do not neglect to draw up a contract, big or small, with the time fixed for paying back the debt |
| ذَٰلِكُمْ أَقْسَطُ عِندَ اللَّهِ وَأَقْوَمُ لِلشَّهَادَةِ وَأَدْنَىٰ أَلَّا تَرْتَابُوا | እንዲህ ማድረጋችሁ አላህ ዘንድ በጣም ትክክል ለምስክርነትም አረጋገጭ ላለመጠራጠራችሁም በጣም ቅርብ ነው | This is more equitable in the eyes of God, and better as evidence and best for avoiding doubt |
| إِلَّا أَن تَكُونَ تِجَارَةً حَاضِرَةً تُدِيرُونَهَا بَيْنَكُمْ فَلَيْسَ عَلَيْكُمْ جُنَاحٌ أَلَّا تَكْتُبُوهَا | ግን በመካከላችሁ እጅ በጅ የምትቀባበሏት ንግድ ብትኾን ባትጽፏት በናንተ ላይ ኃጢአት የለባችሁም | But if it is a deal about some merchandise requiring transaction face to face, there is no harm if no (contract is drawn up) in writing. |
| وَأَشْهِدُوا إِذَا تَبَايَعْتُمْ | በተሻሻጣችሁም ጊዜ አስመስክሩ | Have witnesses to the deal |
| وَلَا يُضَارَّ كَاتِبٌ وَلَا شَهِيدٌ | ጸሐፊም ምስክርም ባለ ጉዳይ ጋር አይጉዳዱ | (and make sure) that the scribe or the witness is not harmed |
| وَإِن تَفْعَلُوا فَإِنَّهُ فُسُوقٌ بِكُمْ | ይህንን ብትሠሩም እርሱ በእናንተ የሚጠጋ አመጽ ነው | If he is, it would surely be sinful on your part |
| وَاتَّقُوا اللَّهَ | አላህንም ፍሩ | And have fear of God |
| وَيُعَلِّمُكُمُ اللَّهُ | አላህም ያሳውቃችኋል | for God gives you knowledge |
| وَاللَّهُ بِكُلِّ شَيْءٍ عَلِيمٌ | አላህም ነገሩን ሁሉ ዐዋቂ ነው | and God is aware of everything |

# 5. AMHARIC-ARABIC NMT

In this work, we adopted openNMT Attention-based Encoder-Decoder architecture, because attention mechanisms are being progressively used to enhance the performance of NMT by selectively focusing on sub-parts of the sentence during translation [24]. As described in [25], "NMT takes a conditional language modeling view of translation by modeling the probability of a target sentence $w_1{:}T$ given a source sentence $x_1{:}S$ as $P(w_t{:}T|x) = \prod_1^t P(w_t|w_{1:t-1}, x)$ . This distribution is estimated using an Attention-based Encoder-Decoder architecture".

Two special kinds of Recurrent Neural Network (RNN) LSTM and GRU which are capable of learning long-term dependencies are used in this research work. RNN is a type of neural network for sequential data that can remember its inputs due to an internal memory which is more suited for machine learning problems. It can produce predictive results in sequential data that the information cycles through a loop when it makes a decision. It takes into consideration the current inputs and also previously received inputs, which is learned earlier [26].

LSTM was first introduced by S. Hochreiter and J. Schmidhuber [27], to avoid the long-term dependency problem. LSTM inherit the exact architecture from standard RNNs, with the exception of the hidden state. The memory in LSTMs (called cells) takes as input the previous state and the current input. Internally, these cells decide what to keep in and what to eliminate from the memory. Then, they combine the previous state, the current memory, and the input. LSTM calculates a hidden state $h_t$ as;

$$i_t = \sigma(x_t U_i + h_{t-1} W_i)$$

$$f_t = \sigma(x_t U_f + h_{t-1} W_f)$$

$$o_t = \sigma(x_t U_o + h_{t-1} W_o)$$

$$\widetilde{C}_t = \tanh(x_t U_{\widetilde{C}} + h_{t-1} W_{\widetilde{C}})$$

$$C_t = \tanh(f_t * C_{t-1} + i_{t-1} * \widetilde{C}_t)$$

$$h_t = \tanh(C_t * o_t)$$

where t, i, f, o, W, U are called the time step, input gate, forget gate, output gate, the recurrent connection at the previous and current hidden layer, and the weight matrix connecting the inputs to the current hidden layers respectively. $\widetilde{C}_t$ is a "candidate" hidden state that is computed based on the current input and the previous hidden state. C is the internal memory of the unit.

GRU extends LSTM with a gating network generating signals that act to control how the present input and previous memory work to update the current activation, and thereby the current network state. Gates are themselves weighted and are selectively updated [28]. For GRU, the hidden state $h_t$ is computed as;

$$Z_t = \sigma(x_t U_Z + h_{t-1} W_Z)$$

$$r_t = \sigma(x_t U_r + h_{t-1} W_r)$$

$$\tilde{h}_t = \tanh(x_t U_h + (r_t * h_{t-1}) W_h)$$

$$h_t = (1 - Z_t) * h_{t-1} + Z_t * \tilde{h}_t$$

where, $\tilde{h}_t$ is activation r is a reset gate, and z is an update gate.

Both LSTM and GRU are designed to resolve the vanishing gradient problem which prevents standard RNNs from learning long-term dependencies through gating mechanism. The general architecture of LSTM and GRU adopted from [29] [30] is shown in Figure 1.

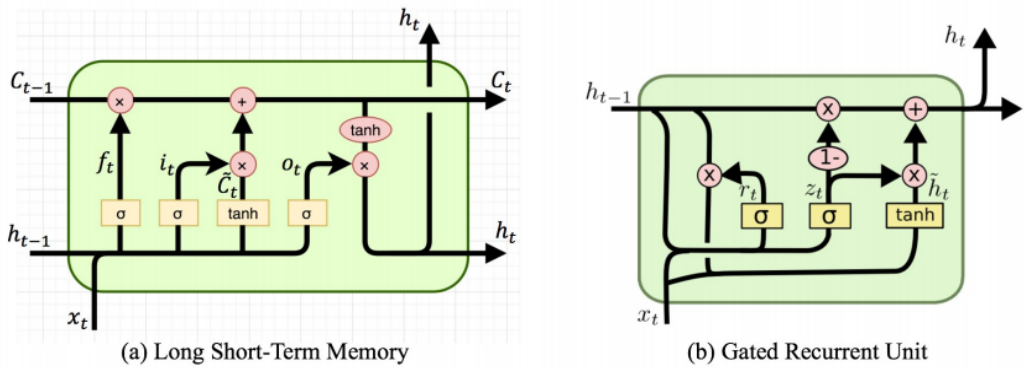

Figure 1: LSTM and GRU Architecture

A basic form of NMT consists of two components; an encoder which computes a representation of source sentence S and a decoder which generates one target word at a time and hence decomposes the conditional probability [24]. The Attention-based Encoder-Decoder architecture used for Amharic-Arabic NMT is shown in Figure 2.

In general, the proposed model works as follows:

1. Reads the input words one by one to obtain a vector representation from every encoder time step using LSTM/GRU based encoder
2. Provide the encoder representation to the decoder
3. Extract the output words one by one using another LSTM/GRU based decoder that is conditioned on the selected inputs from the encoder hidden state of each time step

With this setting, the model is able to selectively focus on useful parts of the input sequence and hence, learn the alignment (matching segments of original text with their corresponding segments of the translation).

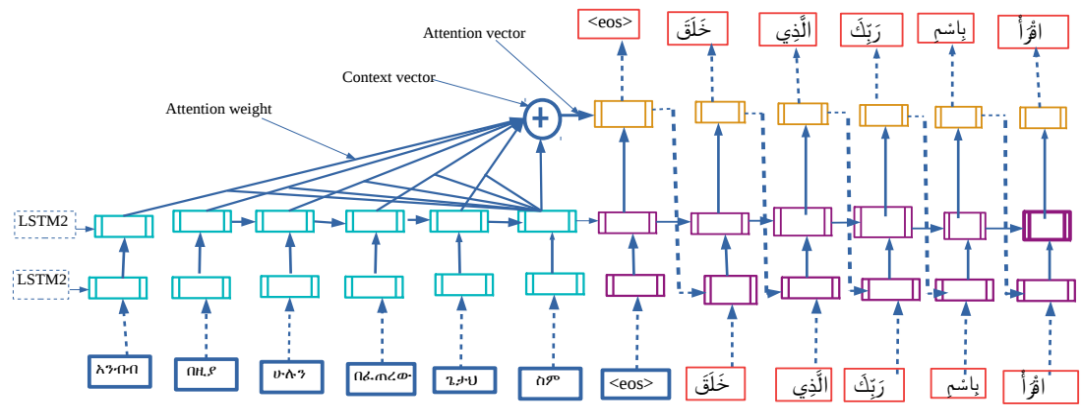

*Figure 2: Attention based Encoder-Decoder architecture for Amharic-Arabic NMT*

As shown in Figure 2, Using LSTM Attention based Encoder-Decoder, each word from the source sentence is associated with a vector $w \epsilon R^d$ and will be transformed into $[w_0, w_1, w_0, w_2, w_3, w_4] \epsilon R^{dx5}$ by the encoder, and then an LSTM over this sequence of vectors is computed. This will be the encoder representation (attention weights) e = $[e_0, e_1, e_0, e_2, e_3, e_4]$. The attention weights and word vectors of each time step is fed to another LSTM cell to compute the context vector which is computed as:

$h_t = LSTM(h_{t-1}, [w_{i-1}, C_t])$

$s_t = g(h_t)$

$p_t = softmax(s_t)$

$i_t = argmax(p_t)$

where, g is a trans-formative function that outputs a vocabulary size vector. A soft-max is then applied to $s_t$ to maximize it to a vector of probability $p_t \epsilon R^V$. Each entry of $p_t$ will measure how likely is each word in the vocabulary and the highest probability $p_t$ is taken as $i_t = argmax(p_t)$, corresponding vector of $w_{i_{t-1}} = w_{i_t}$

$\alpha_{t'} = f(h_{t-1}, e_{t'}) \in R \text{ for all } t'$

$\bar{\alpha} = softmax(\alpha)$

$C_t = \sum_{t_i=0}^{n} \bar{\alpha}_{t'} e_{t'}$

The vector $C_t$ is attention or context vector which is computed at each decoding step first with the function $f(h_{t-1}, e_{t'}) \to \alpha_t' \epsilon R$ then a score for each hidden state $e_t'$ of the encoder is computed. The sequence of $\alpha_t'$ is normalized using a soft-max and $C_t'$ is computed as the weighted average of $e_t'$. The same procedure is also applied for GRU based NMT

## 6. EXPERIMENTS AND RESULTS

An OpenNMT system which is an open-source toolkit for NMT is used to construct the NMT model and translate the text in Amharic to the Arabic language. Preprocessing of both Amharic and Arabic scripts have a great impact on the performance of the NMT system. Sentences are split and aligned manually and then all punctuation marks are removed from texts. After extensive experiments, the maximum source and target sequence length are set to 44, maximum batch size for training and validation is set to 80 and 40 respectively and learning rate to 0.001 with Adam optimization for both LSTM and GRU RNN type. The remaining parameters are used as default. The system saves the model for each of 10,000 training samples and then computes accuracy and perplexity of each model 10 times. Perplexity is a measure of how easily a probability distribution (the model) predicts the next sequence. A low perplexity indicates that the translation model is good at predicting/translating the test set.

$$PP(W) = P(w_1 w_2 ...... w_n)^{-\frac{1}{n}}$$

Table 3 and Table 4 shows LSTM-based and GRU-based NMT evaluation, where, "Val ppl", "Val Acc" "Av. pred score" and "Pred ppl" represents Validation Perplexity, Validation Accuracy, Average Prediction Score and Prediction Perplexity respectively. The result indicate that LSTM-based NMT outperforms GRU-based NMT. Since this is the first experiment done on Amharic and Arabic parallel text corpus we consider it as a good performance with small size corpus.

Table 3. LSTM-based NMT Evaluation

| Epochs | BLEU-Score | Val. PPL | Val. Accuracy | Av. Pred. Score | Pred. ppl |
|---|---|---|---|---|---|
| 1 | 0.11 | 12725 | 33.21 | -0.49 | 1.64 |
| 2 | 0.11 | 41181.5 | 33.50 | -0.40 | 1.49 |
| 3 | 0.11 | 100996 | 33.64 | -0.35 | 1.41 |
| 4 | **0.12** | **100417** | **34.34** | **-0.34** | **1.40** |
| 5 | 0.12 | 99881.1 | 34.32 | -0.34 | 1.40 |
| 6 | 0.12 | 99876.1 | 34.33 | -0.34 | 1.40 |
| 7 | 0.12 | 99876 | 34.33 | -0.34 | 1.40 |
| 8 | 0.12 | 99876 | 34.33 | -0.34 | 1.40 |
| 9 | 0.12 | 99876 | 34.33 | -0.34 | 1.40 |
| 10 | 0.12 | 99876 | 34.33 | -0.34 | 1.4 |

Table 4. GRU-based NMT Evaluation

| Epochs | BLEU-Score | Val. PPL | Val. Accuracy | Av. Pred. Score | Pred. ppl |
|---|---|---|---|---|---|
| 1 | 0.108 | 13647 | 32.65 | -0.51 | 1.66 |
| 2 | 0.101 | 65598.4 | 32.68 | -0.39 | 1.48 |
| 3 | 0.098 | 172950 | 32.38 | -0.35 | 1.43 |
| 4 | 0.105 | 173231 | 33.10 | -0.34 | 1.40 |
| 5 | **0.105** | **175635** | **33.12** | **-0.34** | **1.40** |
| 6 | 0.105 | 175701 | 33.11 | -0.34 | 1.40 |
| 7 | 0.105 | 175702 | 33.11 | -0.34 | 1.40 |
| 8 | 0.105 | 175702 | 33.11 | -0.34 | 1.40 |
| 9 | 0.105 | 175702 | 33.11 | -0.34 | 1.40 |
| 10 | 0.105 | 175702 | 33.11 | -0.34 | 1.40 |

The models are evaluated using Bilingual Evaluation Understudy (BLEU). BLEU is a score for comparing a candidate translation of the text with reference translations. The primary programming task for a BLEU implementer is to compare n-grams of the candidate with the n-grams of the reference translation and count the number of matches. These matches are position-independent. More the matches, better the candidate translation is. BLEU is inexpensive to calculate and it is quick to use. It is expressed as the following equation [31];

$$BLEU = P_B exp\left(\sum_{n=0}^{N} w_n log P_n\right)$$

where $p_n$ is an n-gram precision that uses n-grams up to length N and positive weights $w_n$ that sum to one.

We also compared the two recurrent units LSTM and GRU based OpenNMT translation algorithm with Google Translation System which is a free multilingual translation system developed by Google to translate multilingual text [6] and the results are shown in Figure 3. LSTM based OpenNMT outperforms over GRU based OpenNMT and Google Translation system, which is BLEU score of 12%, 11%, and 6% respectively.

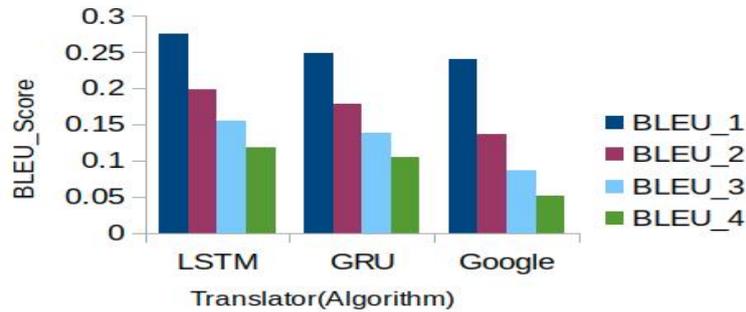

*Figure 3: Best BLEU-Scores of LSTM and GRU based OpenNMT translation and Google Translation System*

## 7. CONCLUSION

Many researchers have investigated to solve translation problems for many language pairs and NMT has recently shown promising results on multiple language pairs. However, much of the research on this area has focused on European languages despite these languages being very rich in resources. Since Amharic and Arabic languages lack parallel corpus for the purpose of developing NMT, small size Amharic-Arabic parallel text corpora have been constructed to train the Amharic to Arabic NMT system by splitting the verses manually into separate sentences of Amharic language as a source sentence and Arabic language as a target sentence. Using the constructed corpus LSTM-based and GRU-based NMT models are developed and evaluated using BLEU. The results are also compared with Google Translation system. Since this is the first experiment done on Amharic and Arabic parallel text corpus, we consider it as a good performance for small size corpus. Extensive experiments with a large amount of training data could be implemented for better performance.

## Authors

Ibrahim Gashaw Kassa is a Ph.D. candidate at Mangalore University Karnataka State, India, since 2016. He graduated in 2006 in Information System from Addis Ababa University, Ethiopia. In 2014, he obtained his master's degree in Information Technology from the University of Gondar, Ethiopia., and he serves as a lecturer at the University of Gondar from 2009 to May 2016. His research interest is in Cross-Language Information Retrieval, Machine translation Artificial Intelligence Natural Language Processing

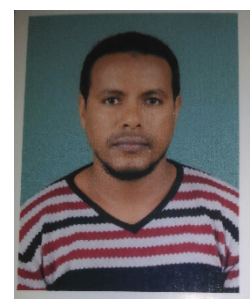

Dr. H L Shashirekha is a Professor in the Department of Computer Science, Mangalore University, Mangalore, Karnataka State, India. She completed her M.Sc. in Computer Science in 1992 and Ph.D. in 2010 from University of Mysore. She is a member of Board of Studies and Board of Examiners (PG) in Computer Science, Mangalore University. She has several papers in International Conferences and published several papers in International Journals and Conference Proceedings. Her area of research includes Text Mining and Natural Language Processing.

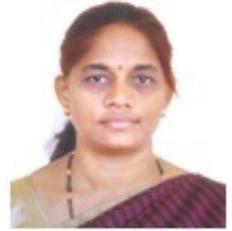